# Building a Kannada POS Tagger Using Machine Learning and Neural Network Models

Ketan Kumar Todi[1], Pruthwik Mishra[2*], and Dipti Misra Sharma[2]

[1] Manipal Institute of Technology
`todiketan1996@gmail.com`
[2] NLP & MT Lab, KCIS, LTRC, IIIT-Hyderabad
`pruthwik.mishra@research.iiit.ac.in,dipti@iiit.ac.in`

**Abstract.** POS Tagging serves as a preliminary task for many NLP applications. Kannada is a relatively poor Indian language with very limited number of quality NLP tools available for use. An accurate and reliable POS Tagger is essential for many NLP tasks like shallow parsing, dependency parsing, sentiment analysis, named entity recognition. We present a statistical POS tagger for Kannada using different machine learning and neural network models. Our Kannada POS tagger outperforms the state-of-the-art Kannada POS tagger by 6%. Our contribution in this paper is three folds - building a generic POS Tagger, comparing the performances of different modeling techniques, exploring the use of character and word embeddings together for Kannada POS Tagging.

## 1 Introduction

Part of Speech (POS) tagging is one of the basic applications of NLP on any language. It is a process of assigning a tag to every word in a sentence. It depends not only on the word whose tag we need to identify, but also on its neighboring words, as the same word can have different roles to play depending on the context. It serves as a preliminary task for carrying out tasks such as chunking, dependency parsing, named entity recognition on any language. So our work focuses on carrying out POS tagging for Kannada. We have tried several different approaches for Part of Speech tagging. The different models used are CRF [16], SVM [8], structured perceptron (StructPercept) [7] and a range of neural network architectures using word embeddings as well as character level embeddings [18] which are explained in section 4. We have used the macro F1-score as an evaluation metric for the accuracy of our approaches.

One of the main problems faced in POS tagging is Unknown or Out-Of-Vocabulary (OOV) words, which can be overcome by using character embeddings. We also use word embedding for the words already present in our vocabulary as it provides us with semantic information which also proves to be useful for our purpose. This paper is organized as the following - section 2 explains the previous work and section 3 details about the corpus and the tagset used. Different approaches employed are explained in the subsequent sections. Results constitutes sections 5 and Error Analysis & Observation are presented in section 6. We conclude our paper with the Conclusion & Future Work section.

---
[*] denotes equal contribution

## 2 Previous Work

POS tagging on Indian languages and especially on Dravidian Languages is quite a difficult task due to the unavailability of annotated data for these languages. Various techniques have been applied for POS tagging in Indian languages. Gadde et al.[11] used morphological features with TNT HMM tagger [5] and obtained 92.36% for Hindi and 91.23 % in Telugu. The Hindi POS tagger used Hindi Treebank [3] of size 450K. Ekbal et al. [10] used SVM for POS tagging in Bengali obtaining 86% accuracy. The POS tagging in morphologically richer Dravidian Indian languages has always posed a great challenge for researchers. Malayalam is a highly agglutinative language in the Dravidian family. Sandhi splitting or word segmentation between conjoined words should precede the POS tagging to find word boundaries. Devadath and Sharma [9] explored the significance of sandhi splitting on shallow parser and built a POS tagger using CRF. Their POS tagger performed well with 90.45% accuracy. Antony et al. [1] used SVM with lexicon to obtain 94% accuracy. A semi-supervised pattern-based bootstrapping technique was implemented by Ganesh and team [12] to build a Tamil POS Tagger. Their system scored 87.74% accuracy on 20000 documents containing 271K unique words.

Very little work has been done on Kannada because of scarcity of quality annotated data. Most of the recent works in POS tagging on Kannada have been experimented only with traditional ML techniques like HMM, CRF or SVM. One of such endeavor was by Shambhavi and Kumar [4] where they showed that their CRF model achieved higher accuracy of 84.58% compared to trigram HMM tagger with 79.9%. They had used the EMILE corpus consisting of mixtures of domains with a training data of size 51K and test set of size 3K. Antony and Soman [2] used a linear SVM with oneVsrest classification to obtain the state-of-the-art accuracy of 86%.

## 3 Corpus Details

We used a corpus which was publicly available [4]. The annotation of the dataset had been funded by DeiTY, Govt. of India. The corpus contained sentences from 3 different domains. The details of the corpus domain wise is given as per table 1. As discussed in section 2 the problem of agglutination, the available corpus is word segmented.

From the available corpus, we had split train and test corpus according to the table 2.

---

[3] ltrc.iiit.ac.in/treebank_H2014
[4] http://ltrc.iiit.ac.in/showfile.php?filename=downloads/kolhi/

Table 1. Domain Wise Corpus Details

| Domain | #Sentences | #Tokens |
|---|---|---|
| General | 8.9K | 109K |
| Tourism | 2K | 24K |
| Conversational | 2.2K | 16K |
| **Total** | **13.1K** | **149K** |

Table 2. Train Test Split

| Type | #Sentences | #Tokens |
|---|---|---|
| Train | 10.5K | 119K |
| Test | 1.6K | 30K |
| **Total** | **13.1K** | **149K** |

The corpus was useful for designing a generic POS Tagger because of its size and diversity in terms of domains.

**Tagset** Most of the cited work in the previous section used a smaller tagset of 27-30 tags with a limited corpus size. The tagset used for this task is a unified XML based POS tagset called BIS [5] (Bureau of Indian Standards) tagset. The pos schema is based on W3C XML Internalization best practices. This tagset contains 40 POS tags for Kannada. But the datasets used in this paper had only 37 of those.

## 4 Approaches

We mainly used two kinds of approaches:

- Machine Learning (ML) Approaches
- Neural Networks

The approaches are detailed in the below subsections.

### 4.1 ML Approaches

Two different machine learning techniques were used for this task - CRF and SVM. For CRF implementation, we used the open source CRF++ toolkit [6].

**Features Used** The features used in CRF, SVM and structured perceptron. We used the word itself, prefixes upto length 3, and suffixes upto length 4. An additional binary feature was also used which checked the length of a word. If the word length is greater than 3, then the feature is set as 'MORE', 'LESS'

---
[5] http://tdil-dc.in/tdildcMain/articles/134692Draft%20POS%20Tag%20standard.pdf
[6] https://taku910.github.io/crfpp/

otherwise. Context window of different lengths were also taken as features to check its impact on POS tagging. For this paper, the context windows used were of size 3 and 5. The context window of size 3 was referred to as [-1, +1] and size 5 as [-2, +2]. We also used word bigrams and trigrams as features. For the purpose of POS tagging using CRF, we tried a number of experiments as mentioned in the table 5. An example for the features for a kannada word "ಅಷ್ಟೇ" ("Aṣṭē", English gloss - "that's it") is shown below in table 3

**Table 3.** Features for Kannada

| Feature | Example |
|---|---|
| word | ಅಷ್ಟೇ |
| Prefix length 1 | ಅ |
| Prefix length 2 | ಅಷ |
| Prefix length 3 | ಅಷ್ |
| Suffix length 1 | ೇ |
| Suffix length 2 | ಟೇ |
| Suffix length 3 | ್ಟೇ |
| Suffix length 4 | ಷ್ಟೇ |
| Length | MORE |

We trained a SVM model using multiple binary classifiers using the yamcha [7] SVM model. The features were the prefix, suffix features of all the words in a context window of size 5 and the POS tag output of the SVM model for the last two words.

### 4.2 Neural Networks

We tried a range of neural network and deep learning approaches starting with structured perceptron where we used the same set of features as used in CRF and the results obtained were comparable to that of CRF. Structured perceptron was implemented using the seqlearn [8] library. All the recurrent neural network models were built using Keras [6] deep learning framework. Adam [15] optimizer was used for word embeddings experiments with a batch size of 32 and RM-SPROP [19] for joint character and word embedding experiments and batch size of 64. The networks were trained for 25 epochs to learn the parameters of the respective models. For all the recurrent networks, the loss function used was sparse categorical cross entropy. While training the networks, we used 10% of the training data for validation.

As POS tagging is sequence labeling task, we modeled it as a sequence-to-sequence learner. We started with the Vanilla RNN network, which gave an output POS tag for every input word. The vanilla learning model where the lengths of input and output sequences are same is the perfect architecture for POS Tagging. We used just the word as an input and passed the entire sentence to the neural network and we used the FastText [3] Kannada word embeddings and used a random initialization for unknown words. We used different recurrent architectures LSTM and bi-directional LSTM (biLSTM) [13]. These all experiments were carried out with the help of pre-trained Kannada embeddings. We tried to implement the encoder-decoder architecture where the entire sequence of the words or a sentence was represented as a single vector. This vector was then passed onto a decoder which gave an output POS tag for every input word. The encoder-decoder architecture was not performing as expected, so we did not report the results in this paper.

---

[7] http://chasen.org/ taku/software/yamcha/
[8] https://github.com/larsmans/seqlearn

Table 4. OOV words statistics in Train Test sets

| Type  | #Tokens | #OOV-Words |
|-------|---------|------------|
| Train | 119K    | 14.7K      |
| Test  | 29K     | 3.7K       |

The word vectors of many words were not available because the word embedding model was trained on a limited wikipedia dump for Kannada. The details of OOVs in train and test datasets are given

To overcome the problem of OOV words, we switched to character embedding. The character embeddings were learned with the help of a biLSTM network. The output of the biLSTM was the representation of the word by using the character embeddings of individual letters present inside the word. This helped in improving the accuracy as it provided a better way of handling OOVs. This was applied with the final layer being recurrent neural networks which outputted a sequence of POS tags. Different variants of recurrent networks SimpleRNN, LSTM [14] and Bi-LSTM were experimented with a vanilla architecture. The architecture of the neural network used for learning character embeddings and its subsequent learning for generating POS tags are shown in figures 1 and 2.

We finally combined the word embedding with the character embedding where a better syntactic representation was learned to gain character level information for a morphologically rich language like Kannada. This approach was suggested by Ling et al [18]. This was similar to finding the prefix and suffixes in the ML models. All the above models were also tried with another setting in which the word embeddings were not frozen and was allowed to be learned by the neural network. The main motivation behind this was that word embeddings learned for specific purposes are better than the ones learned using context information for other purposes. Though it did not improve the accuracy a lot, but had more data been available then we are of the view that the accuracy would have improved a lot more. We also tried to form the character embedding using CNN [17] for POS tagging, but the results were not motivating. One of the main rea-sons might be the lack of enough data, for the CNN to learn proper character embeddings.

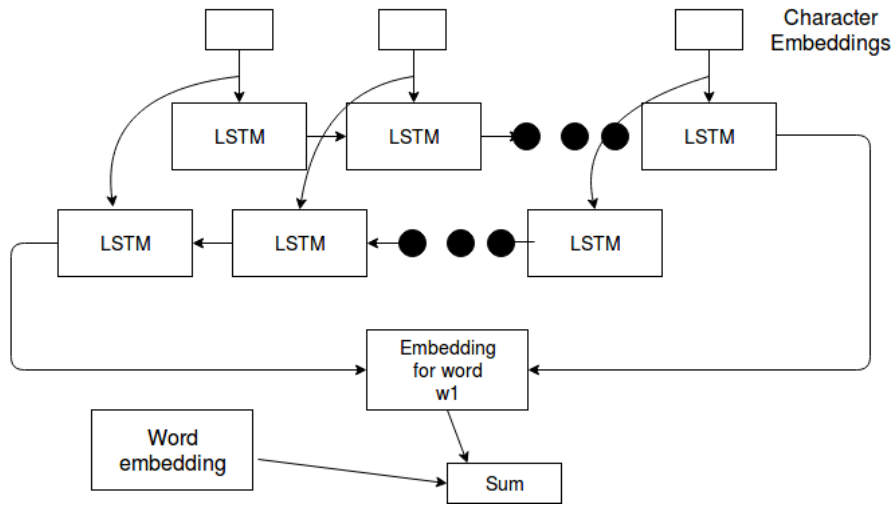

**Fig. 1.** Character Embedding and Final Word Embedding Learning

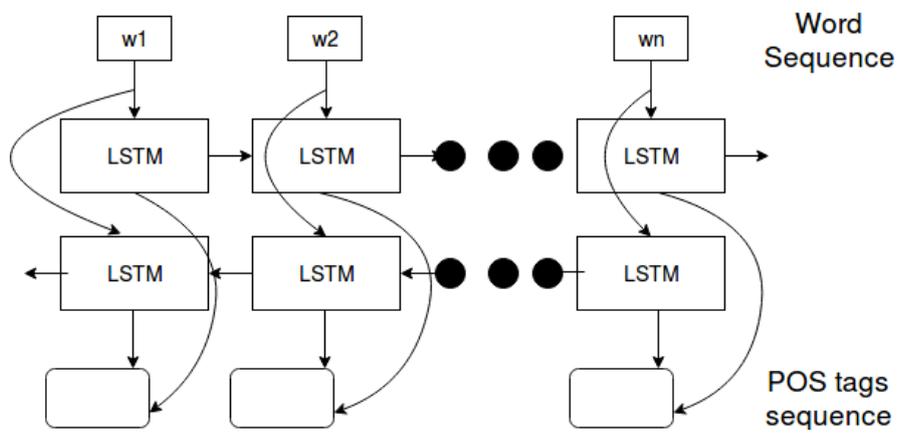

**Fig. 2.** Sequence To Sequence Learning using BiLSTM for POS Tagging

Table 5. Precision, Recall and F1-scores for ML Models

| Model | Features | P | R | F |
|---|---|---|---|---|
| CRF | word features | 0.91 | 0.91 | 0.91 |
| | word features + [-1, +1] | **0.92** | **0.92** | **0.92** |
| | word features + [-2, +2] | **0.92** | **0.92** | **0.92** |
| | word features + [-1, +1] + word bigrams | **0.92** | **0.92** | **0.92** |
| | word features + [-1, +1] + word trigrams | **0.92** | **0.92** | **0.92** |
| SVM | word features + [-2, +2] | 0.91 | 0.91 | 0.90 |
| StructPercept | word features | 0.91 | 0.91 | 0.91 |
| | word features + [-1, +1] | 0.91 | 0.91 | 0.91 |
| | word features + [-2, +2] | 0.91 | 0.91 | 0.91 |
| | word features + [-1, +1] + word bigrams | 0.91 | 0.91 | 0.91 |
| | word features + [-1, +1] + word trigrams | 0.91 | 0.91 | 0.91 |

## 5 Experimental Results

We report the experimental results in 2 different tables. Table 5 shows the results of different models where feature engineering was required. Table 6 reports the accuracies of neural network models which used (pre-trained) word embedding and/or character embedding. We report the performance in terms of macro precision, recall and F1-scores. Macro F1-scores averages the F1-scores for individual POS classes. All the previous systems used overall percentage accuracy metric to evaluate their system. Our best performing CRF system gave 92.31% overall accuracy.

Table 6. Precision, Recall and F1-scores for Neural Models

| Model | Features | P | R | F |
|---|---|---|---|---|
| SimpleRNN | non pre-trained word embeddings | 0.84 | 0.84 | 0.84 |
| LSTM | | 0.86 | 0.85 | 0.85 |
| biLSTM | | 0.87 | 0.87 | 0.87 |
| SimpleRNN | pre-trained word embeddings | 0.85 | 0.85 | 0.85 |
| LSTM | | 0.87 | 0.87 | 0.87 |
| biLSTM | | 0.88 | 0.87 | 0.88 |
| SimpleRNN | char + word embeddings | 0.89 | 0.89 | 0.89 |
| LSTM | | 0.90 | 0.91 | 0.91 |
| biLSTM | | **0.92** | **0.92** | **0.92** |

## 6 Error Analysis and Observation

The sources of major errors are listed as follows:

– Errors in the Gold Data
– Ambiguities between Nouns and Adjective
– Ambiguities between Particle and Conjunction
– Ambiguities between Verbs and Auxiliary Verbs
– Ambiguities between Quantitatives and Adjectives
– Ambiguities between Adverbs and Adjectives

Confusion matrix revealed some more ambiguities in predictions like ambiguities between Finite Verbs and Common Nouns, ambiguities between common nouns and adverbs. We could observe that some tokens training data were inconsistently tagged. For example, the token "," should have been tagged as "RD_PUNC". Out of 3689 occurrence of the token in the training corpus, 68 times it had been tagged as "RD_SYM" and "RD_PUNC" remaining 3621 times. The tagger was robust to the noise which could be seen from the F1-score of the "RD_PUNC" being 0.99. We also noticed that due to finer categories of verbs, a non finite verb (V_VM_VNF) could easily be tagged as gerund (V_VM_VNG) because of the morphological similarities. We observed that the pre-trained word embeddings models outperformed the non pre-trained counterparts. This might be due to the sparse nature of embeddings in case of non pre-trained embeddings. When the pre-trained word embedding were coupled with the compositional character representation, it improved the performance of the system by at least 2%. For CRF and structured perceptron, giving additional features of bigrams and trigrams did not improve the system. Window size of 3 was found to be sufficient neighborhood for POS tagging. We could see that all the sequence learners perform equally well when considerable size of training data was provided.

## 7 Conclusion and Future Work

The best F1-score of 0.92 was obtained in the CRF model on using a window feature of [-2,+2] and in the neural network model where we used the character embedding along with pre-trained word embeddings. The accuracy of the neural networks can be improved by training it on a larger dataset, and also the word embeddings used can be trained on a larger dataset, so as to reduce the OOV words. The character and word embedding model used by us is not specific for Kannada and can be used for other languages, especially for agglutinative language like Malayalam as it would benefit a lot from the character embeddings model used. The neural network model used in this paper can also be used for carrying out chunking if annotated data is available .Chunking models tend to give better accuracy when the correct word POS is available. We could explore attention based neural networks for learning the best POS sequence for a sentence.

## Acknowledgement

We would like to thank KCIS, LTRC, MIT, Manipal for making the kannada corpus available for researchers. We thank TDIL, DeiTy, Govt Of India for taking the initiative for creating resources for Indian languages and releasing them for the NLP community.